\documentclass[11pt,english]{article}
\usepackage{times}
\usepackage[font=small,labelfont=bf,tableposition=top]{caption}

\usepackage[nottoc,numbib]{tocbibind}
\usepackage{tikz-dependency}
\usepackage{nicefrac}
\usepackage{framed}
\usepackage{url}
\usepackage{marginnote}
\usepackage{pdfpages}

\usepackage{babel}

\usepackage{xspace}
\usepackage{wrapfig}
\usepackage{sectsty}
\usepackage{graphicx}
\usepackage{amsfonts,eucal,amsbsy,amsopn,amsmath}
\usepackage[T1]{fontenc}

\usepackage[hmargin=1in, vmargin=1in]{geometry}
\usepackage{enumitem}
\usepackage{parskip}
\usepackage{natbib}

\bibpunct{(}{)}{;}{a}{,}{,}

\setitemize{noitemsep,topsep=10pt,parsep=0pt,partopsep=0pt}
\setenumerate{noitemsep,topsep=10pt,parsep=0pt,partopsep=0pt}

\title{Annotating Character Relationships in Literary Texts}

\author{
  Philip Massey and Patrick Xia\\
  School of Computer Science\\
  Carnegie Mellon University\\
  \url{{pmassey,paxia}@cmu.edu}
  \and
  David Bamman\\
  School of Information\\
  University of California, Berkeley\\
  \url{dbamman@berkeley.edu}
  \vspace{15pt}
    \and
  Noah A.~Smith\\
  Computer Science \& Engineering\\
  University of Washington\\
  \url{nasmith@cs.washington.edu}
}

\date{}

\begin{document}
\maketitle

\section{Overview}

We present a dataset of manually annotated relationships between characters in literary texts, in order to support the training and evaluation of automatic methods for relation type prediction in this domain \citep{DBLP:journals/corr/MakazhanovBK14,Kokkinakis2013} and the broader computational analysis of literary character \citep{Elson2010,bammanliterary2014,vala-EtAl:2015:EMNLP,flekova-gurevych:2015:EMNLP}.  In this work, we solicit annotations from workers on Amazon Mechanical Turk on four dimensions of interest: for a given pair of characters, we collect judgments as to the \textbf{coarse-grained category} (professional, social, familial), \textbf{fine-grained category} (friend, lover, parent, rival, employer), and \textbf{affinity} (positive, negative, neutral) that describes their primary relationship in a text.   We do not assume that this relationship is static; we also collect judgments as to whether it \textbf{changes} at any point in the course of the text.  

The annotations describe character dyads in 109 texts ranging from Homer's \emph{Iliad} to Joyce's \emph{Ulysses} (see section \ref{corpus} for a full list).  Rather than relying on annotators' expertise in these texts, we frame the annotation problem as one of estimating the relationship as depicted in a third-party summary (SparkNotes); this allows annotators to provide judgments on the relationship between pairs of characters by only reading a summary of a book, and not the book itself.  While this approach naturally loses the nuance of a truly expert opinion, it allows us to broadly characterize a large number literary dyads, assess the feasibility of this annotation strategy, and provide a foundation on which other work can build. All data is openly available at \url{http://github.com/dbamman/characterRelations}, and we encourage contributions and corrections.

\section{Data Collection}

Our primary corpus selection criterion was that a text is both available on Project Gutenberg (to enable computational analysis of an open-access text) and is the subject of a study guide on \url{www.sparknotes.com} (to enable annotation by non-experts).  SparkNotes provides a detailed summary of the plot and major characters in texts, often a structured format (e.g., a section denoted ``Character List'' ); from this summary we extract all characters and use them to populate the following questionnaire.

\clearpage
\hrule
\vspace{10pt}
\textbf{Fictional Character Relationship Analysis}

In this task, you'll be identifying the relationship type that exists between two characters in \emph{The Good Soldier} by Ford Madox Ford, using a description and summary of that work from SparkNotes. (For example, if reading \emph{To Kill a Mockingbird}, you'd mark that Atticus Finch is the father of Scout Finch.)

We expect this task to take approximately 20 minutes.
Please read the "Character List," "Plot Overview," and "Character Analysis / Analysis of Major Characters" pages here: \url{http://www.sparknotes.com/lit/goodsoldier/}.

After reading these pages, list all of the relationships that you can identify between the characters described there. To complete a relationship, find the two characters in the dropdown menus below with First Character and Second Character. Then, answer the following questions using these guidelines:

\emph{Affinity}:

How do the two characters feel toward each other? For example, if they are friendly, select "Positive". If it is unclear how they feel toward each other or if they do not have strong opinions about each other, select "Neutral". If they are enemies or rivals, or hate each other, select "Negative".

\emph{Category}:

How are the two characters related? If they are friends, select "Social". If they share a relationship because they work with each other, select "Professional". If they are family, then select "Familial".

\emph{Kind}:

Specifically, how is the First Character related to the Second Character? For example, if the First Character is the husband of the Second Character, then select "husband". Please keep the Category and the Kind consistent:

If you selected "Social", then the Kind must be one of these:
\begin{itemize}
\item friend
\item enemy
\item acquaintance
\item lovers
\item unrequited love interest (X is in love with Y, but Y is not in love with X)
\item rivals
\end{itemize}
If you selected "Professional", then the Kind must be one of these:
\begin{itemize}
\item employer
\item employee
\item colleague
\item servant
\item master
\item student
\item teacher
\item client
\item person offering service to client (e.g., lawyer)
\end{itemize}

If you selected "Familial", then the Kind must be one of these:
\begin{itemize}
\item husband/wife
\item brother/sister
\item cousin
\item uncle/aunt
\item niece/nephew
\item child
\item parent
\item grandchild
\item grandparent
\item orphan
\item foster parent
\item step-child
\item step-parent
\item in-law relation (e.g., mother-in-law; specify in detail)
\item half relation (e.g., half-sister; specify in detail)
\end{itemize}

\emph{Change}:

Does the relationship between the two characters significantly change at some point in the book?  For example, does a "positive" relationship become "negative", or do "lovers" become "husband/wife" or "friends" become "enemies"?  If so, select "Yes" here and describe the change in the Detail section below. If the relationship remains the same throughout (for example, "brother/sister"), then select "No".

\emph{Detail}:

If the relationship type you feel holds between two characters was not provided, or you want to provide additional information, enter it here.

If two characters hold multiple relations to each other (such as "Professional: colleague" and "Social: lovers"), enter those multiple relations in different rows below. Please identify a total of 10 relations between characters in the rows below.

(Users then select a pair of characters from the following list and label the relations specified above):

\begin{itemize}
\item John Dowell
\item Florence Hurlbird Dowell
\item Leonora Powys Ashburnham
\item Captain Edward Ashburnham
\item Nancy Rufford
\item Jimmy
\item Uncle John Hurlbird
\item Maisie Maidan
\item Rodney Bayham
\item Mrs. Basil
\item La Dolciquita
\item The Misses Hurlbird
\item Selmes
\item Major Rufford
\end{itemize}

\hrule
\vspace{10pt}

We present this questionnaire to workers on Amazon Mechanical Turk, soliciting two independent judgments for each of the 109 literary texts.  Since we are soliciting judgments regarding any 10 character pairs in the text (and not a fixed set of such dyads), many of the character pairs from different annotators for the same work do not overlap.  We collect a total of 2,170 annotations; among these, 392 character dyads have annotations by two different annotators, from which we can calculate agreement statistics.  Table \ref{iaa} lists the inter-annotator agreement rate (and Fleiss' $\kappa$, correcting for chance) for each of the four annotation classes.  The agreement rates for both the coarse- and fine-grained categories are both high, even when correcting for chance ($\kappa = 0.812$ and $0.744$, respectively).  18.1\% of character pairs are judged to exhibit some change over the course of the text.  While annotators display high agreement on this (75.7\%), their agreement is in fact quite low when correcting for chance ($\kappa = 0.208$).  Judging whether a dyad's relationship is \emph{positive, negative} or \emph{neutral} also proves to be quite difficult, with low agreement rates across annotators.

\begin{figure*}[h]
\begin{centering}
\includegraphics[scale=.6]{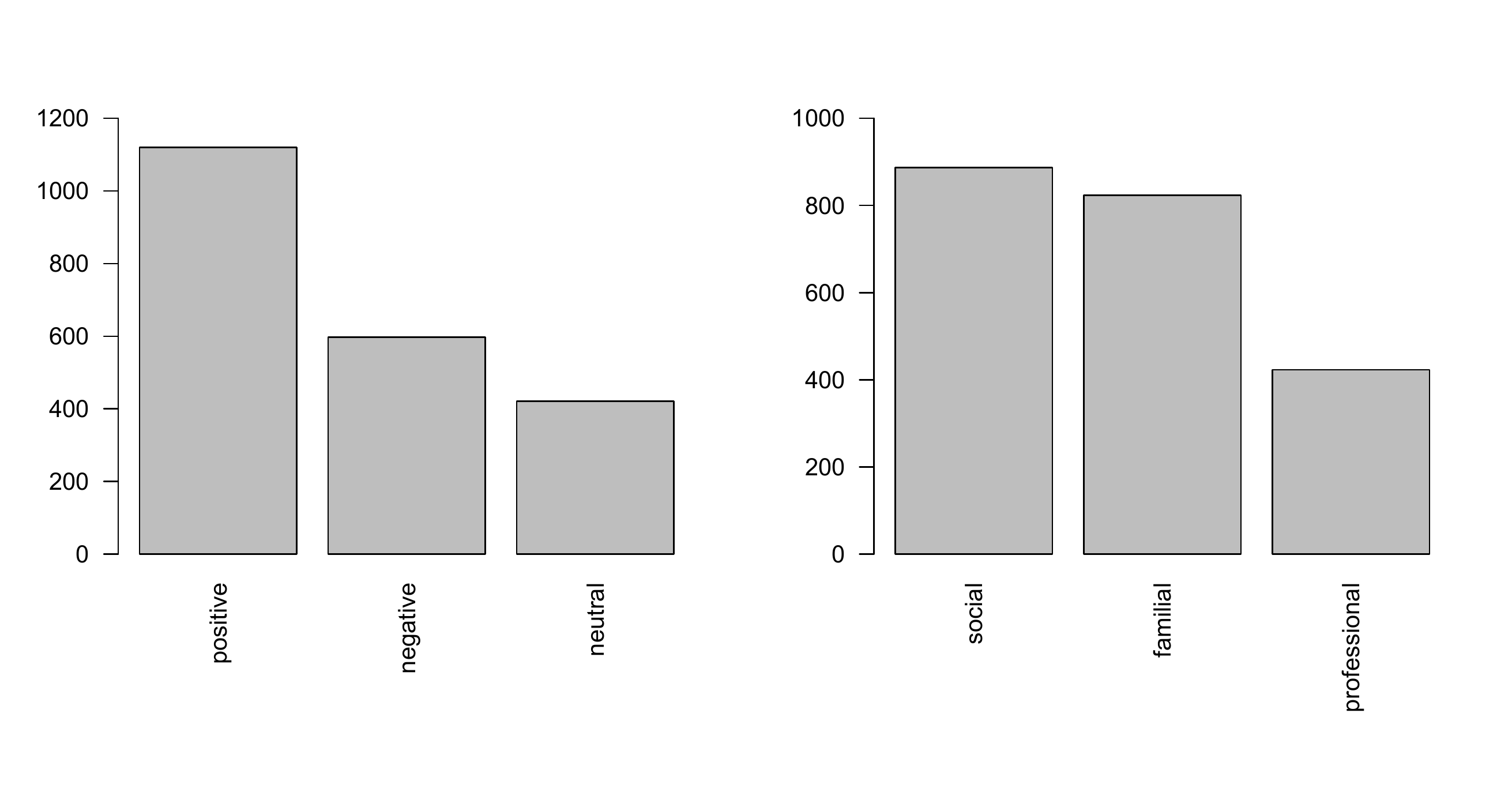}
\caption{\label{cataff} Distribution of affinity annotations (left) and coarse-grained category annotations (right).}
\end{centering}
\end{figure*}

\begin{figure*}[h]
\begin{centering}
\includegraphics[scale=.7]{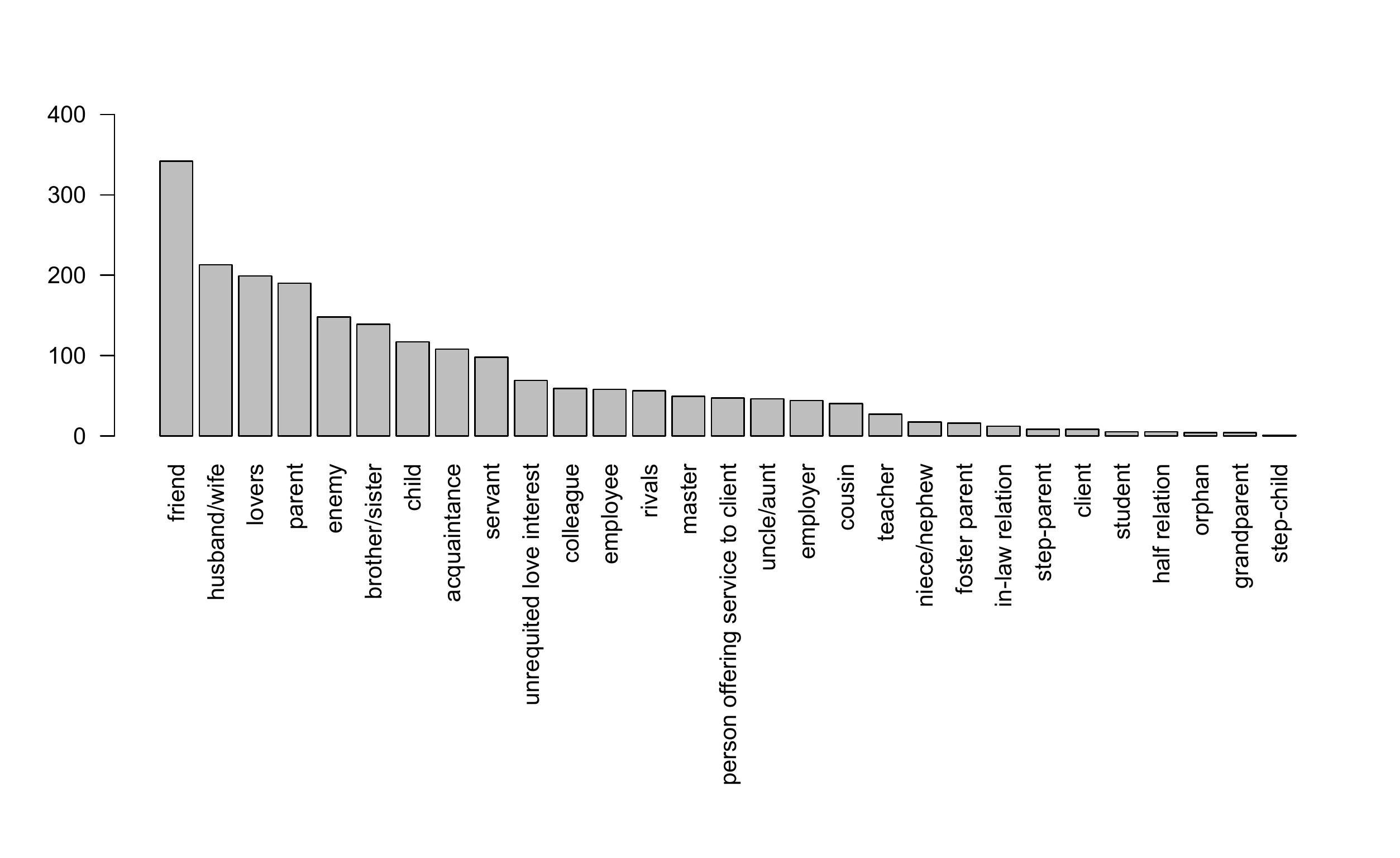}
\caption{\label{kind} Distribution of fine-grained category annotations.}
\end{centering}
\end{figure*}

Figure \ref{cataff} shows the distribution of affinity and coarse-grained category annotations, while figure \ref{kind} shows the distribution of fine-grained category annotations.

\begin{table}[ht!]
\begin{center}
\footnotesize
\begin{tabular}{| l | c | c | c |} \hline
Label type&IAA&Fleiss' $\kappa$&$n$ \\ \hline \hline
Affinity&0.627&0.364&391 \\ \hline
Category&0.879&0.812&389 \\ \hline
Kind&0.765&0.744&392\\ \hline
Change&0.757&0.208&371 \\ \hline
\end{tabular}
\end{center}
\caption{\label{iaa}Inter-annotator agreement (IAA) rates, along with chance-corrected Fleiss' $\kappa$ for the four annotation tasks. $n$ denotes the sample size of directed character pairs with two annotations; sample size exhibits question-level variability due to incomplete responses.}\end{table}

\section{Texts}\label{corpus}

\begin{itemize}
\item Alexandre Dumas. The Count of Monte Cristo, The Three Musketeers
\item Aristophanes. Lysistrata
\item Bram Stoker. Dracula
\item Charles Dickens. A Tale of Two Cities, Bleak House, Great Expectations, Hard Times, Oliver Twist
\item Charlotte Perkins Gilman. Herland
\item Christopher Marlowe. The Jew of Malta
\item E. M. Forster. Howards End
\item Edith Wharton. Ethan Frome, The House of Mirth
\item Edmond Rostand. Cyrano de Bergerac
\item F. Scott Fitzgerald. This Side of Paradise
\item Frances Hodgson Burnett. The Secret Garden
\item Franz Kafka. The Trial
\item Frederick Douglass. Narrative of the Life of Frederick Douglass
\item George Bernard Shaw. Major Barbara, Pygmalion
\item George Eliot. Adam Bede, Middlemarch, Silas Marner
\item Gustave Flaubert. Madame Bovary
\item Harriet Beecher Stowe. Uncle Tom's Cabin
\item Henrik Ibsen. A Doll's House, Ghosts, Hedda Gabler
\item Henry James. The American, The Portrait of a Lady, The Turn of the Screw
\item Herman Melville. Typee
\item Hermann Hesse. Siddhartha
\item Homer. The Iliad
\item Jack London. White Fang
\item James Fenimore Cooper. The Last of the Mohicans
\item James Joyce. A Portrait of the Artist as a Young Man, Dubliners, Ulysses
\item Jane Austen. Emma, Mansfield Park, Northanger Abbey, Persuasion, Pride and Prejudice, Sense and Sensibility
\item John Milton. Paradise Lost
\item Joseph Conrad. Heart of Darkness, Lord Jim
\item L. M. Montgomery. Anne of Green Gables
\item Leo Tolstoy. Anna Karenina, War and Peace
\item Louisa May Alcott. Little Women
\item Marcel Proust. Swann's Way
\item Mark Twain. The Adventures of Tom Sawyer
\item Miguel de Cervantes. Don Quixote
\item Nathaniel Hawthorne. The House of the Seven Gables, The Scarlet Letter
\item Oscar Wilde. An Ideal Husband, The Picture of Dorian Gray
\item Plato. The Republic
\item Robert Louis Stevenson. Dr. Jekyll and Mr. Hyde, Kidnapped, Treasure Island
\item Sinclair Lewis. Babbitt, Main Street
\item Stephen Crane. Maggie: A Girl of the Streets
\item Theodore Dreiser. Sister Carrie
\item Thomas Hardy. Far from the Madding Crowd, Jude the Obscure, The Mayor of Casterbridge, The Return of the Native
\item Thomas Kyd. Spanish Tragedy
\item Upton Sinclair. The Jungle
\item Virgil. The Aeneid
\item Voltaire. Candide
\item Willa Cather. O Pioneers!
\item William Shakespeare. A Midsummer Night's Dream, Antony and Cleopatra, As You Like It, Hamlet, Henry IV Part 1, Henry IV Part 2, Henry V, Henry VI Part 1, Henry VI Part 2, Henry VI Part 3, Henry VIII, Julius Caesar, King John, King Lear, Love's Labours Lost, Macbeth, Measure for Measure, Much Ado About Nothing, Othello, Richard II, Romeo and Juliet, The Comedy of Errors, The Merchant of Venice, The Merry Wives of Windsor, The Taming of the Shrew, The Tempest, The Two Gentlemen of Verona, The Winter's Tale, Timon of Athens, Titus Andronicus, Troilus and Cressida, Twelfth Night
\end{itemize}

\section{Data}

All annotations are freely available at \url{http://github.com/dbamman/characterRelations}.  Rather than reconciling disagreements between annotators or filtering out incomplete responses, we are making all collected data available; each annotations is paired with an (anonymized) identifier of the annotator who provided it.

\section{Acknowledgments}

We thank Matthew Jockers, Andrew Piper, Ted Underwood and Chris Warren for helpful feedback.  This research was supported by a Google research award and was made possible through the use of computing resources made available by the Open Science Data Cloud (OSDC), an Open Cloud Consortium (OCC)-sponsored project.

\small

\bibliographystyle{plainnat}

\bibliography{characterRelations}

\end{document}